\documentclass{bmvc2k}
\usepackage{graphicx}
\usepackage{bm}
\usepackage{graphics}

\usepackage{pifont}
\title{Open Logo Detection Challenge}

\addauthor{Hang Su}{http://www.eecs.qmul.ac.uk/~hs308/}{1}
\addauthor{Xiatian Zhu}{eddy@visionsemantics.com}{2}
\addauthor{Shaogang Gong}{https://www.eecs.qmul.ac.uk/~sgg/}{1}

\addinstitution{
 School of EECS\\
 Queen Mary University of London\\
 London E1 4NS, UK
}
\addinstitution{
 Vision Semantics Limited\\
 327 Mile End Road\\
 London E1 4NS, UK
}

\runninghead{H. Su, X. Zhu, S. Gong}{Open Logo Detection Challenge}

\begin{document}

\maketitle

\begin{abstract}
Existing logo detection benchmarks consider 
artificial deployment scenarios by assuming
that large training data with fine-grained bounding box annotations 
for each class are available for model training.
Such assumptions are often invalid in realistic logo detection scenarios where 
new logo classes come progressively and require to be detected
with little or none budget for exhaustively labelling fine-grained training data 
for every new class.
Existing benchmarks are thus
unable to evaluate the true performance of a logo detection method 
in realistic and open deployments.
In this work, we introduce a more realistic
and challenging logo detection setting, called {\em Open Logo Detection}. 
Specifically,
this new setting assumes fine-grained labelling only on 
a small proportion of logo classes whilst the remaining classes
have no labelled training data to simulate the
open deployment.
We further create an open logo detection benchmark, called {\em QMUL-OpenLogo},
to promote the investigation of this new challenge.
QMUL-OpenLogo contains 27,083 images from 352 logo classes,
built by aggregating/refining 7 existing datasets
and establishing an open logo detection evaluation protocol.
To address this challenge, we propose a {\em Context Adversarial Learning} (CAL)
approach to synthesising training data with coherent logo instance appearance
against diverse background context for enabling more effective optimisation of contemporary 
deep learning detection models.
Experiments show the performance advantage of CAL 
over existing state-of-the-art alternative methods
on the more realistic and challenging open logo benchmark.
The QMUL-OpenLogo benchmark is publicly available at
\url{https://qmul-openlogo.github.io/}.
\end{abstract}


\section{Introduction}
\label{sec:intro}

Logo detection in unconstrained scene images
is crucial for a variety of real-world vision applications, 
such as brand trend prediction for commercial research and
vehicle logo recognition for intelligent transportation
\cite{romberg2011scalable,romberg2013bundle,pan2013vehicle}.
It is inherently a challenging task 
due to the presence of varying sized logo instances  
in arbitrary scenes with uncontrolled illumination,
multi-resolution, occlusion and background clutter (Fig. \ref{fig:dataset} (c)).
Existing logo detection methods typically  
consider a small number of logo classes
with the need for large sized training data annotated with object bounding boxes
\cite{joly2009logo,kalantidis2011scalable,romberg2011scalable,revaud2012correlation,romberg2013bundle,boia2014local,li2014logo,pan2013vehicle,liao2017mutual}.
Whilst this controlled setting allows for a straightforward adoption of the state-of-the-art general object detection models \cite{ren2015faster,girshick2015fast,redmon2017yolo9000},
it is 
unscalable to dynamic real-world logo detection applications
where more new logo classes become of interest
during model deployment,
with the availability of
only their clean design images (Fig. \ref{fig:dataset}(a)).
To satisfy such incremental demands, 
prior methods are significantly limited
by the extremely high cost needed for labelling 
a large set of per-class training logo images \cite{russakovsky2015imagenet}.
Whilst this requirement is significant for practical deployments, 
it is ignored in existing logo detection benchmarks
which consider only the unscalable fully supervised learning evaluations.

This work considers the realistic and challenging open-ended logo detection challenge. 
To that end, we introduce a new \textbf{Open Logo Detection} problem, where we have 
limited fine-grained object bounding box annotation
in real scene images
for only a small proportion of logo classes (supervised)
with the remaining classes (the majority)
totally unlabelled (unsupervised).
As logo is a visual symbol, we have the clean logo designs for all target classes (Fig. \ref{fig:dataset}(a)).
The {\em objective} is to establish a logo detection model
for all logo classes by exploiting the small labelled training set
and logo design images in a scalable manner.
One approach to open logo detection is by jointly learning logo detection and classification as YOLO9000 \cite{redmon2017yolo9000} so that
the model can learn to detect logo objects from
the labelled training images while 
learn to classify all logo design images.
This method relies on robust object detection generalisation
learned from labelled classes to unlabelled classes, and 
rich appearance diversity of object instances. 
Both assumption are invalid in our setting. 
Another alternative approach is synthesising training data \cite{su2016deep}
which overlays logo designs with geometry/illumination
variations into context images
at random scales and locations.
%
But, it introduces {\em appearance inconsistency} 
between logo instances and scene context (Fig. \ref{fig:synthetic_images}(a)), 
which may impede the model generalisation.
%

In this work, we address the open logo detection challenge
by presenting a Context Adversarial Learning (CAL) approach to automatically generate
context consistent synthetic training data.
Specifically, the CAL takes as input artificial images with
superimposed logo designs \cite{su2016deep}, and outputs corresponding
images with context consistent logo instances.
This is a pixel prediction process, which
we formulate as an image-to-image translation problem
in the Generative Adversarial Network framework \cite{goodfellow2014generative}.

Our contributions are: 
{\bf (1)} We scale up logo detection to 
dynamic real-world applications without fine-grained
labelled training data for newly coming logo classes 
and present a novel {\em Open Logo Detection} setting.
This differs significantly from existing fully supervised logo detection problems 
with the exhaustive need for object instance box labelling for all classes 
and hence having poor deployment scalability in reality.
To our knowledge, this is the first attempt of investigating
such a scalable logo detection scenario in the literature.
{\bf (2)} We introduce a new QMUL-OpenLogo benchmark
for providing a standard test of open logo detection
and facilitating a like-for-like comparative evaluation
in the future studies.
QMUL-OpenLogo is created based on 7 publicly available logo detection datasets
through a careful logo classes merging and filtering process
along with a benchmarking evaluation protocol.
{\bf (3)} We propose a Context Adversarial Learning (CAL) approach
to synthesising context coherent training data for enabling effective 
learning of state-of-the-art object detection models in order to tackle
the open logo detection challenge in a scalable manner. 
Importantly, CAL requires no exhaustive human labelling therefore generally applicable to 
any unsupervised logo classes.
Experiments show the performance
advantage of CAL for open logo detection
on the QMUL-OpenLogo benchmark in comparison to state-of-the-art
approaches YOLO9000 \cite{redmon2017yolo9000}
and Synthetic Context Logo (SCL) \cite{su2016deep}
in contemporary object detection frameworks.

\section{Related Works}

\noindent \textbf{Logo Detection}
Traditional methods for logo detection rely on hand-crafted features
and sliding window based localisation
\cite{li2014logo,revaud2012correlation,romberg2013bundle,boia2014local,kalantidis2011scalable}.
Recently, deep learning methods 
\cite{iandola2015deeplogo,hoi2015logo,su2016deep,su2017weblogo,liao2017mutual} have been proposed
which use generic object detection models \cite{girshick2014rich,ren2015faster,girshick2015fast,redmon2017yolo9000}.
However, 
these methods are not scalable to realistic 
large deployments due to the need for:
(1) Accurately labelled training data per logo class;
(2) Strong object-level bounding box annotations.
%
%
One exception is \cite{su2017weblogo,su2018scalable} where noisy
web logo images are exploited without manual labelling of 
object instance boxes.
This method exploits a huge quantity of data to mine 
sufficient correct logo images,
and is restricted for non-popular and new brand logos 
which may lack web data.
%
%
Moreover, all the above-mentioned methods
assume the availability of real training images for ensuring model
generalisation. 
This further reduces their scalability and usability
in real-world scenarios when
many logo classes have no training
images from real scenes such as those newly introduced logos.  
In this work, we investigate this under-studied {\em Open Logo Detection} setting,
where the majority of logo classes 
have no training data.
%

\vspace{0.1cm}
\noindent \textbf{Synthetic Data}
There are previous attempts to exploit synthetic data for training deep CNN models. 
Peng et al. \cite{peng2015learning} used 3D CAD object models 
to generate 2D images by varying the projections and orientations to augment 
the training data in few-shot learning scenarios.
This method is based on the R-CNN model \cite{girshick2014rich} 
with the proposal generation component independent from fine-tuning the classifier, 
making the correlation between objects and background context suboptimal. 
The work of \cite{su2015render} used synthetic data rendered from 3D models against varying background 
to enhance the training images of a pose model. 
Su et al. \cite{su2016deep} similarly generated synthetic images by
overlaying logo instances with appearance changes 
on random background images.
Rather than randomly placing exemplar objects \cite{su2015render,su2016deep,peng2015learning}, 
Georgakis et al. \cite{georgakis2017synthesizing} 
performed object-scene compositing based on accurate scene segmentation,
similar as \cite{gupta2016synthetic} for text localisation.
%
%
%
These existing works mostly aim to generate images with varying object appearance.
%
In contrast, we consider the consistency between objects and the surrounding context
for generating appearance coherent synthetic images.
%
Conceptually, our method is complementary to the aforementioned approaches
when applied concurrently.




\section{QMUL-OpenLogo: Open Logo Detection Benchmark}
\label{sec:benchmark}
For enabling open logo detection performance test,
we need to establish a corresponding benchmark which
the literature lacks. 
To that end, it is necessary to collect a large number of logo classes
for simulating the real-world deployments at scales.
Given the tedious process of logo class selection,
image data collection and filtering, as well as 
fine-grained bounding box annotation \cite{su2012crowdsourcing},
we propose to re-exploit the existing logo detection datasets.
%

\begin{table}  
	\centering
	\setlength{\tabcolsep}{0.1cm}
	\caption{
		Statistics of logo detection datasets
		used for constructing the QMUL-OpenLogo benchmark.
		Scale is the ratio 
		of the logo instance area to the whole image area.
	}
	\vskip -0.0cm
	\label{tab:dataset}
	\resizebox{\columnwidth}{!}
	{
	\begin{tabular}{l||c|c|c|c}
		\hline
		Dataset & Logos & Images & min$\sim$max (mean) Instances / Class & min$\sim$max (mean) Scale (\%) \\
		\hline \hline
		FlickrLogos-27 \cite{kalantidis2011scalable}
		& 27 & 810 & 35$\sim$213 (80.52) 
		&  0.0160$\sim$100.0 (19.56) \\
		\hline
		FlickrLogos-32 \cite{romberg2011scalable}
		& 32 & 2,240 & 73$\sim$204 (106.38)
		&  0.0200$\sim$99.09 (9.16)  \\
		\hline
		Logo32plus \cite{bianco2017deep} 
		& 32 & 7,830 &  132$\sim$576 (338.06)
		& 0.0190$\sim$100.0 (4.51)	 \\ 
		\hline
		BelgaLogos \cite{joly2009logo}  
		& 37 & 1,321  & 2$\sim$223 (57.08)
		& 0.0230$\sim$69.04 (0.91)  \\
		\hline
		WebLogo-2M(Test) \cite{su2017weblogo}
		& 194 & 4,318 & 18$\sim$204 (40.63)
		& 0.0180$\sim$99.67 (7.69) \\
		\hline 
		Logo-In-The-Wild \cite{tuzko2017open}
		& 1196 & 9,393 & 1$\sim$1080 (23.49)
		& 0.0007$\sim$95.91 (1.80) \\
		\hline 
		SportsLogo \cite{liao2017mutual}
		& 20 & 1,978 & 108$\sim$292 (152.25) 
		& 0.0100$\sim$99.41 (9.89) \\
		\hline \hline
		\bf QMUL-OpenLogo
		& 352 
		&27,083 & 10$\sim$1,902 (88.25)
		& 0.0014$\sim$100.0 (6.09)  \\
		\hline 
	\end{tabular}
  	}
	\vspace{-0.2cm}
\end{table}

\vspace{0.1cm}
\noindent {\bf Source data selection} 
To maximise the context richness of logo images,
we assemble 7 existing publicly accessible logo detection datasets (Table \ref{tab:dataset}) sourced from diverse domains
to establish the QMUL-OpenLogo evaluation benchmark. 
All these datasets together present significant 
logo variations and therefore represent the truthful logo detection challenge
as encountered in real-world unconstrained deployments.
We only used the test data of WebLogo-2M
\cite{su2017weblogo} since its training data are
noisy without labelled object bounding boxes which are
required for model performance evaluation.

\vspace{0.1cm}
\noindent {\bf Logo annotation and class refinement} 
We need to make logo class definition consistent in QMUL-OpenLogo
provided that different definitions exist across datasets.
In particular, \texttt{\small Logo-In-The-Wild} (LITW) treats different logo variations of the same brand as distinct logo classes.
For example, Adidas trefoil/text are treated
as two different classes in LITW
but as one class in all other datasets. 
We adopted the latter more common criterion
by merging all fine-grained same-brand logo classes 
from LITW.
We 
combined all logo image data of the same logo class from all selected datasets.
We also cleaned up erroneous annotations by
removing those with the size of bounding box exceeds the whole image size
and/or obviously wrong box coordinates (e.g. $x_\text{min}$ > $x_\text{max}$).
To ensure that each logo class has sufficient test data,
we further removed those extremely small classes with less than 10 images.
Moreover, we manually verified 1$\sim$3 random images per class
and filtered out those classes with incorrect labels on selected images.
These refinements result in a QMUL-OpenLogo dataset with 
27,189 images of 309 logo classes (Table \ref{tab:dataset}).

\begin{table}  
	\centering
	\setlength{\tabcolsep}{0.4cm}
	\caption{
		Train/val/test data split in
		the QMUL-OpenLogo benchmark.
	}
	\label{tab:split}
	\resizebox{\columnwidth}{!}
	{
		\begin{tabular}{l||c|c|c|c|c}
			\hline
			Type & Classes & Train Img & Val Img & Test Img & Logo Design Img \\
			\hline \hline
			Supervised 
			& 32 & 1,280 & 1,019 & 9,168 & 32 (1 per class)
			\\
			\hline
			Unsupervised 
			& 320 & 0 & 1,562 & 14,054 & 320 (1 per class)
			\\ 
			\hline \hline
			Total 
			& 352 & 1,280 & 2,581 &  23,222 & 352 (1 per class)
			\\
			\hline 
		\end{tabular}
	}
\end{table}

\begin{figure}
	\centering
	\bmvaHangBox{
		\includegraphics[width=\linewidth]{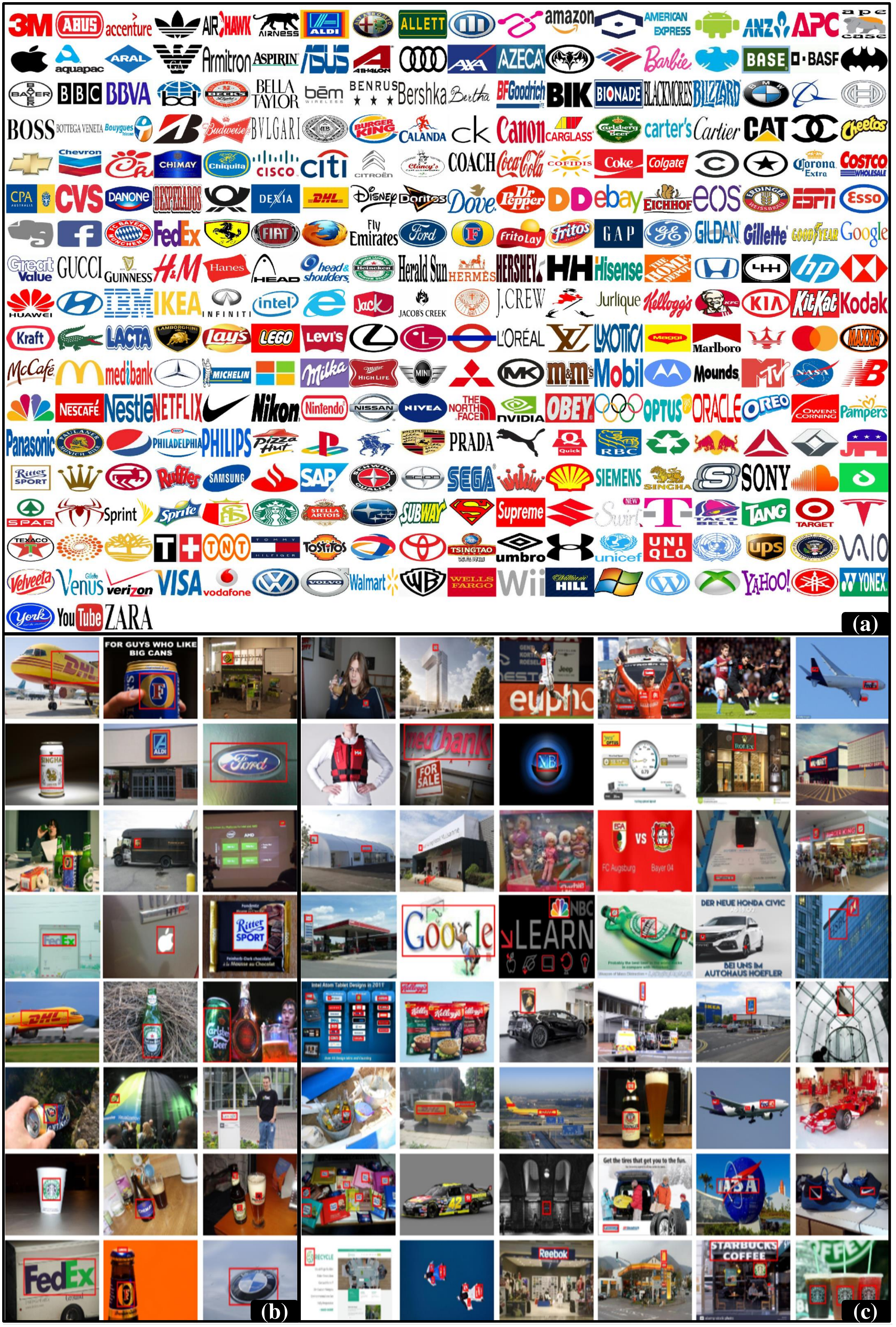}
	}
	\caption{
		Example images of the QMUL-OpenLogo benchmark.
		(a) Clean logo design images;
		(b) Training images; 
		(c) Test images.
	}
	\label{fig:dataset}
\end{figure}

\vspace{0.1cm}
\noindent {\bf Train/val/test data partition} 
For model training and evaluation on the QMUL-OpenLogo dataset as a test benchmark,
we standardise the train/val/test sets in the following two steps:
(1) We split all logo classes into two disjoint groups:
one is {\em supervised} with labelled bounding boxes in real images, 
and the other is {\em unsupervised}, i.e. the open logo detection setting.
In particular, we selected all 32 classes in the popular FlickrLogo32 dataset \cite{romberg2011scalable} as supervised ones whilst the remaining 277 classes as unsupervised.  
(2) For each supervised class, 
we assigned the original \texttt{\small trainval} set (40 images per class) of FlickrLogo32 as the \texttt{\small train} data.
For open logo detection, no real training images are 
available for unsupervised classes.
Among the remaining images, 
we further performed a random 10\%/90\% split for the \texttt{\small val}/\texttt{\small test} partition on each class.
The data split is summarised in Table \ref{tab:split}.

\vspace{0.1cm}
\noindent {\bf Logo design images}
Similar to \cite{su2016deep}, 
we obtained the clean logo design images
from the Google Image Search by querying the corresponding logo class name.
These images define the logo detection tasks,
one for each logo class (Fig. \ref{fig:dataset} (a)).

\vspace{0.1cm}
\noindent {\bf Benchmark properties}
The QMUL-OpenLogo benchmark has three characteristics:
(1) Highly imbalanced logo frequency distribution
({``Instances/Class''} in Table \ref{tab:dataset});
(2) Significant logo scale variation
({``Scale''} in Table \ref{tab:dataset});
(3) Rich scene context
(Fig. \ref{fig:dataset}(c)).
These factors are essential for creating a benchmark entailing
the true performance test of logo detection algorithms
in real-world large scale deployments.

\section{Synthesising Images by Context Adversarial Learning}

In open logo detection, there are training images $\mathcal{D}^s$ only for a small number of supervised classes $\mathcal{C}^s$,
whilst no training data for other unsupervised classes
$\mathcal{C}^u$ (Table \ref{tab:split}).
To enable state-of-the-art detection 
models \cite{redmon2017yolo9000,ren2015faster,liu2015ssd},
we exploit the potential of synthetic training data.
To this end, we propose a {\bf Context Adversarial Learning} (CAL) approach for rendering logo instance against scene context and improving the appearance consistency.

\subsection{Context Adversarial Learning}

\begin{figure}
	\centering
	\bmvaHangBox{
		\includegraphics[width=\linewidth]{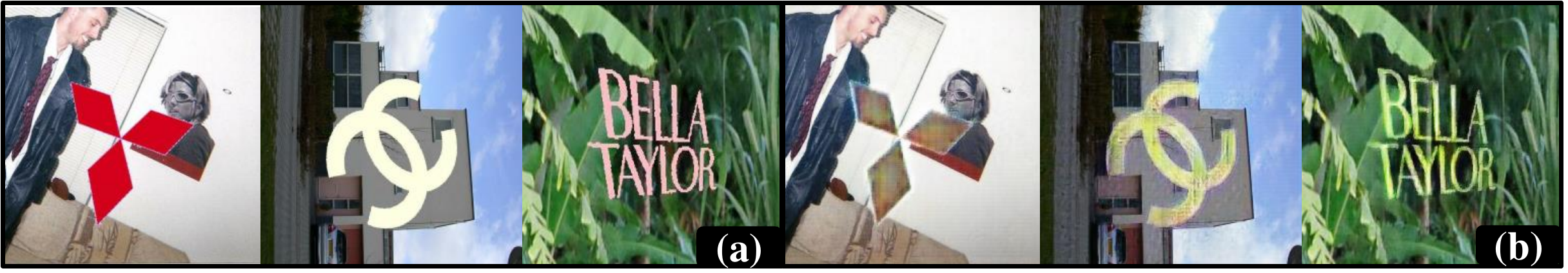}
	}
	\caption{
		Examples of synthetic logo images by (a) SCL \cite{su2016deep}
		and (b) our CAL.
	}
	\label{fig:synthetic_images}
\end{figure}

The proposed CAL takes as input initial synthetic images with
logo objects to generate the corresponding {\em context consistent} synthetic images. These output images serve as 
additional training data for enhancing state-of-the-art detection model generalisation
on real-world unconstrained images. 
CAL is therefore a {\em detection} data augmentation
strategy 
with focus on logo context optimisation.
Conceptually, it is totally complementary 
to other existing data augmentation methods
widely adopted for training classification and detection
deep learning models,
e.g. flipping, rotation, random noise, scaling
\cite{ren2015faster,krizhevsky2012imagenet,redmon2017yolo9000,liu2015ssd}.
In this study, we adopt the SCL \cite{su2016deep}
to generate the initial synthetic data,
i.e. superimpose the logo design images with
spatial and colour transformation
into any given natural scene images (Fig. \ref{fig:synthetic_images}(a)).

\vspace{0.1cm}
\noindent {\bf Model Formulation}
We consider the CAL as an image-to-image ``translation'' problem,
i.e. translating one representation $x$ of a logo scene image into another $y$ of the same content at the pixel level. 
We particularly focus on rendering the logo objects
to be more consistent with the scene context.
Recently, deep neural networks
have been verified as strong models capable of learning to minimise a given objective loss function \cite{goodfellow2016deep,lecun2015deep}.
A straightforward solution may be the common convolutional neural networks (CNNs) which can be supervised to
minimise the Euclidean distance between 
the predicted and ground truth pixel values.
However, such modelling may lead to blurring results
provided that the objective loss is minimised by
averaging all plausible outputs
\cite{pathak2016context,isola2017image}.
How to generate realistic images, the core of CAL, 
remains a generally unsolved problem for CNNs.

Interestingly, this task is exactly the formulation purpose of
the recently proposed Generative Adversarial Networks (GANs)
\cite{goodfellow2014generative,radford2015unsupervised,denton2015deep,salimans2016improved} -- making the generated images 
indistinguishable from realistic ones.
Unlike the manually designed loss functions in CNNs,
%
a GAN model automatically learns a loss that tries to classify if an output image is real (e.g. context consistent) or fake
(e.g. context inconsistent), while simultaneously training a generative model to minimise this loss.
Blurry logo images are clearly fake and therefore 
well suppressed.
Given the dependence on initial synthetic image in our context, 
we explore the image-conditioned GAN
which learns a conditional generative model \cite{goodfellow2014generative,isola2017image}.
Formally, the objective value function of a conditional GAN can be written as:
\begin{equation}
\mathcal{L}_\text{cGAN} (G, D) = \mathbf{E}_{x,y}[\log D(x,y)] + 
\mathbf{E}_{x,z}[\log\big(1-D(x, G(x,z))\big)]
\label{eq:GAN_obj}
\end{equation}
where the generator $G$ tries to minimise this objective value against an adversarial discriminator $D$ which instead tries to maximise the value.
Inspired by the modelling benefits from combining the GAN objective
with pixel distance loss \cite{isola2017image,pathak2016context}, we enhance the conditional adversarial loss (Eq. \eqref{eq:GAN_obj}) with an $L_1$ loss to further 
suppress the blurring possibility:
\begin{equation}
\mathcal{L}_\text{cGAN} (G, D) = \mathbf{E}_{x,y}[\log D(x,y)] + 
\mathbf{E}_{x,z}[\log\big(1-D(x, G(x,z))\big)]
+ \lambda \mathbf{E}_{x,y,z}{\| y - G(x,z) \|_1}
\label{eq:GAN_obj_L1}
\end{equation}
where $\lambda$ controls the weight of the $L_1$ pixel matching loss.
We empirically set $\lambda=100$ in our experiments.
As such, the generator learning is also tied with
the task of being close to the ground truth output $y$
in addition to fooling the discriminator,
whilst 
the discriminator learning remains unchanged.
The optimal solution is:
\begin{equation}
G^* = \arg \; \min_G \; \max_D \mathcal{L}_\text{cGAN} (G, D)
\label{eq:GAN_opt}
\end{equation}
%
The noise $z$ input aims to learn mapping from a distribution 
(e.g. Gaussian \cite{wang2016generative})
to the target domain $y$.
However, this strategy is often ineffective
to capture the stochasticity of conditional distributions
with the noise largely neglected \cite{isola2017image,mathieu2015deep}.
Fortunately, it is not highly necessary to fully model this 
distribution mapping in our problem due to the presence of potentially infinite synthetic images by SCL.
More specifically, 
the variation of $y$ can be more easily captured
by sampling the input synthetic logo images,
than
modelling the entropy of the conditional distributions
through learning a mapping from one distribution to another.
Without $z$,
the model still learns a mapping from $x$ to $y$ in a deterministic manner.

\vspace{0.1cm}
\noindent {\bf Network Architecture}
We adopt the same generator and discriminator architectures as \cite{isola2017image}.
Specifically, the generator is an encoder-decoder network in a U-Net architecture
with the encoder being a 8-layer CNN net
whilst the decoder with a minor structure. 
The discriminator is a 4-layers CNN net.
All convolution layers use 4$\times$4 filters with stride 2.

\begin{figure} [h]
	\centering
	\bmvaHangBox{
		\includegraphics[width=1\linewidth]{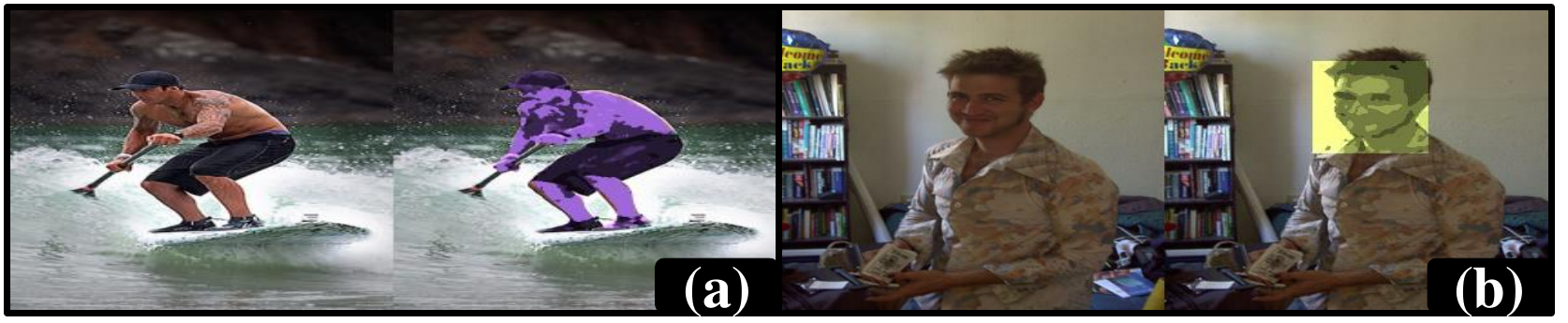}}
	\caption{
		Example CAL training pairs
		generated based on
		(a) object masks of COCO images
		and 
		(b) random regions of non-logo images.
	}
	\label{fig:training pair}
\end{figure}

\vspace{0.1cm}
\noindent {\bf Training Images}
To train the CAL model,
we need a set of training image pairs.
For generalising the model
to rendering the synthetic images by SCL \cite{su2016deep} at test time,
we also apply the SCL to automatically build the training data.
Specifically, given any natural image $\bm{I}$,
we select a region (either a random rectangle or object foreground)
and render it by SCL transformation including image sharpening, median filtering, random colour changes and colour reduction.
This results in a CAL training image pair ($\bm{I}$, $\bm{I}_\text{SCL}$)
where $\bm{I}_\text{SCL}$ is the image with inconsistent object region.
We select two image sources for enhancing context richness:
(1) Non-logo background images from FlickrLogo32 \cite{romberg2011scalable}
on which we use random rectangle regions to generate training pairs;
and
(2) MS COCO images \cite{lin2014microsoft} 
on which we utilise the object masks to 
make training pairs.
Importantly, this method requires no additional labelling in creating 
training data. A number of examples are given in
Fig. \ref{fig:training pair}.

\vspace{0.1cm}
\noindent {\bf Model Optimisation and Deployment}
Given the training data, 
we adopt the standard optimisation approach as \cite{isola2017image}
to train the CAL: we alternate between 
optimising the discriminator $D$, and then 
the generator $G$ in each mini-batch SGD.
We train $G$ to minimise $\log D(x, G(x, z))$
rather than $\log(1-D(x, G(x, z))$ as suggested in \cite{goodfellow2014generative}.
We slow down the learning of $D$ by applying half gradient.
To focus the CAL model on context inconsistent region,
we additionally input the region mask together with $\bm{I}_\text{SCL}$
in another channel, leading to a 4-channels input.
Once the CAL is trained, it can be used to perform
context rendering on the SCL synthesised images.
Example CAL synthesised images are shown
in Fig. \ref{fig:synthetic_images} (b).

\subsection{Multi-Class Logo Detection Model Training}

Given both SCL and CAL synthesised images
and real (supervised classes only) images,
we train a pre-selected deep learning detection model
\cite{ren2015faster,redmon2017yolo9000}.
First training on synthetic data 
and then fine-tuning on real images
may make the detection model biased towards
supervised logo classes whilst significantly
hurting the performance on other unsupervised classes,
as we will show in the experiments (Table \ref{tab:syn_real}).

\section{Experiments}

\noindent {\bf Competitors}
We compared our CAL approach with
two state-of-the-art approaches allowing for
open logo detection:
{\bf (1)} 
SCL \cite{su2016deep}:
A state-of-the-art method for generating synthetic
detection training data with random logo design transformation
and unconstrained background context.
This enables the exploitation of state-of-the-art 
detection models same as the CAL.
We selected two strong deep learning detectors:
Faster R-CNN \cite{ren2015faster}
and 
YOLOv2 \cite{redmon2017yolo9000}.
{\bf (2)}
YOLO9000 \cite{redmon2017yolo9000}: 
A state-of-the-art deep learning detection model based on 
the YOLO architecture \cite{redmon2015you}. 
This model is designed particularly to scale up the detector to
a large number of classes without exhaustive
object instance box labelling. The key idea is
by jointly learning the model from
both bounding box labelled training data of supervised classes and 
image level classification training data of all classes using
mixture mini-batches.  
We adopt the softmax cross-entropy loss for classification
rather than the hierarchy aware loss as used in \cite{redmon2017yolo9000}, 
since there is no semantic hierarchy on logo classes.
%
To improve the model classification robustness against 
uncontrolled context, 
we further performed context augmentation on logo design images (Fig. \ref{fig:dataset}(a))
using the SCL method \cite{su2016deep}.
%

\vspace{0.1cm}
\noindent {\bf Performance Metric}
For the performance measure of logo detection, 
we adopted the standard Average Precision (AP) for each individual logo class,
and the mean Average Precision (mAP) for all classes \cite{everingham2010pascal}.
A detection is considered to be correct when the Intersection over Union (IoU) between the predicted and ground-truth exceeds $50\%$.

\vspace{0.1cm}
\noindent {\bf Implementation Details}
For CAL model optimisation, we adopted the Adam solver \cite{kingma2014adam}
at the learning rate of 0.0002 and momentum parameters 
$\beta_1=0.5$ and $\beta_2=0.999$.
For SCL \cite{su2016deep} and CAL, we generated
100 synthetic images per logo class (in total 35,200).
%

\begin{table}
	\centering
	\caption{Evaluation on the QMUL-OpenLogo benchmark.
		Uns Class: Unsupervised Class;
		Sup Class: Supervised Class.
		Metric: mAP (\%).
		FR-CNN: Faster R-CNN.
		Abs/Rel Gain: the absolute/relative performance gain of CAL over SCL.
		{\bf {\color{red} Red}}: the best results.
	}
	\label{tab:main}
	\setlength{\tabcolsep}{0.15cm}
	\vspace{0.1cm}
	\resizebox{\columnwidth}{!}
	{%
		\begin{tabular}{l||c||c|c||c|c}
			\hline
			Method & All Class & Uns Class & Sup Class & Big Logo & Small Logo  \\
			\hline\hline
			YOLO9000\cite{redmon2017yolo9000} 
			& 4.19 
			& 1.98 & 26.33 & 6.23
			&  1.15 \\
			\hline
			
			YOLOv2\cite{redmon2017yolo9000}+SCL\cite{su2016deep} 
			& 12.10 
			& 8.75 & 45.58 
			& 17.66 & 5.92\\
			YOLOv2\cite{redmon2017yolo9000}+{\bf CAL} 
			& \color{red} \bf 13.14
			& \color{red} \bf 9.55 &\bf 49.17 
			&\color{red}\bf 18.25 &\bf 6.29\\
			\hline
			Abs/Rel Gain (\%) & 1.04/8.60 & 0.80/9.14 & {\bf 3.59}/7.88 & 0.59/3.34 & 0.37/{\bf 6.25} \\
			\hline 
			\hline
			
			FR-CNN\cite{ren2015faster}+SCL\cite{su2016deep} 
			& 12.35
			& 8.51 & 50.74 & 16.94 & 7.87
			\\
			FR-CNN\cite{ren2015faster}+{\bf CAL} &\bf 13.13
			&\bf 9.34 &\color{red} \bf 51.03 &\bf 17.68 &\color{red} \bf 8.69
			\\
			\hline 
			Abs/Rel Gain (\%) 
			& 0.78/6.32 & {\bf 0.83}/9.75 & 0.29/0.57 & 0.74/4.37
			& 0.82/{\bf 10.41} \\
			\hline
		\end{tabular}
	}
\end{table}

\vspace{0.1cm}
\noindent{\bf Comparative Evaluations}
The comparative results on the QMUL-OpenLogo benchmark
are shown in Table \ref{tab:main}.
To look into the detailed performance, we further evaluate 
the performance on unsupervised and supervised logo classes,
as well as big (46.7\%) and small (53.3\%) logo instances split with a threshold of 0.02 scale ratio.
We have these observations:
{\bf (1)}
All methods produce rather poor results ($<14\%$ mAP) on
the QMUL-OpenLogo benchmark,
suggesting that the scalability of current solutions
remains unsatisfied in open logo detection deployments 
with the need for further investigation. 
{\bf (2)}
YOLO9000 \cite{redmon2017yolo9000} 
yields the weakest performance among all methods.
This suggests that joint learning of object classification
and detection in a single loss formulation is ineffective
to solve this challenging problem,
particularly when the classification training data (clean logo design images) are 
limited in appearance variations.
It is also found that such modelling can negatively
affect the performance on supervised classes
with fine-grained labelled training data. 
{\bf (3)}
With CAL, YOLOv2 achieves the best logo detection performance.
While the absolute gain of CAL over the state-of-the-art data synthesising method SCL
is small, larger relative gains are achieved
using either YOLOv2 or Faster R-CNN.
{\bf (4)}
The accuracy on supervised logo classes 
is much better than that on unsupervised ones.
This indicates the high reliance on the manually labelled training 
data for existing state-of-the-art detection models,
and the unsolved challenge of learning from auto-generated synthetic images.
{\bf (5)}
Small logos benefit the largest relative gain from CAL.
This is reasonable because, given limited appearance details 
of small instances, the external contextual information becomes
more important for achieving accurate localisation and recognition. 

To further justify the weak performance of state-of-the-art methods on
the new QMUL-OpenLogo challenge,
we evaluated Faster R-CNN+CAL on the most popular benchmark FlickrLogos-32 \cite{romberg2011scalable}.
We obtained 74.9\% mAP, which closely matches the 73.3\% mAP of \cite{iandola2015deeplogo}
similarly using a deep CNN model.
%
%
%

\begin{figure}[h]
	\centering
	\bmvaHangBox{
		\includegraphics[width=1\linewidth]{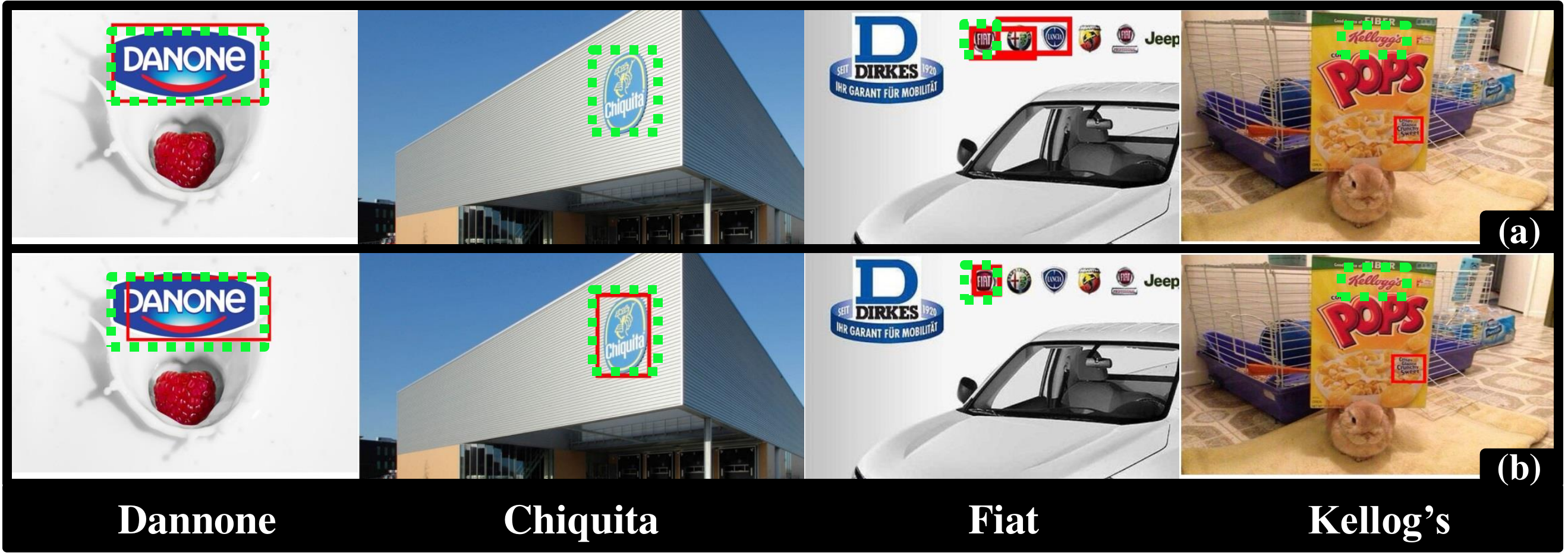}
	}
	\caption{
		Example detection results by (a) SCL \cite{su2016deep} and (b) CAL. 
		Green dash boxes: ground-truth;
		Red solid boxes: model detection.
	}
	\label{fig:model_compare}
\end{figure}

\vspace{0.1cm}
\noindent{\bf Qualitative Examination}
Fig \ref{fig:model_compare} shows four test examples 
by SCL and CAL based on Faster R-CNN.
For big logo instances of ``Danone'' in clean background, 
both models succeed. 
For moderate ``Chiquita'' with viewpoint distortion and small ``Fiat''
with subtle appearance, 
the SCL model fails while CAL remains successful. 
For small ``Kellogs'' instance against complex background clutter, 
both model fail.

\vspace{0.1cm}
\noindent{\bf Further Analysis}
In Table \ref{tab:syn_real}, we evaluated the performance of Faster R-CNN
(1) trained on the mixture of synthetic and real data, or
(2) trained firstly on synthetic data and then fine-tuned real data. 
It is evident that the former produces better performance
except on supervised classes.
This is as expected since the latter will bias the model towards
supervised logo classes in the fine-tuning stage whilst
largely degrading the generalisation capability on unsupervised classes.


\begin{table} [h]
	\centering
	\setlength{\tabcolsep}{0.3cm}
	\caption{Comparing the training methods of detection model (Faster R-CNN)
		using synthetic (syn) and real training data on the QMUL-OpenLogo benchmark.
		Uns Class: Unsupervised Class;
		Sup Class: Supervised Class.
		Metric: mAP (\%).
	}
	\label{tab:syn_real}
	\vspace{0.05cm}
	\resizebox{\columnwidth}{!}
	{%
		\begin{tabular}{l||c||c|c||c|c}
			\hline
			Method & All Class & Uns Class & Sup Class & Big Logo & Small Logo   \\
			\hline\hline
			syn+real 
			& \bf 12.35
			& \bf 8.51 & 50.74 & \bf 16.94 &\bf 7.87 \\
			\hline
			
			syn$\rightarrow$real
			& 6.62 & 1.54 & \bf 57.39 & 8.19 & 3.77\\
			\hline
		\end{tabular}
	}
\end{table}

\vspace{0.1cm}
\noindent{\bf Fully Supervised Learning Evaluation} 
To more extensively evaluate the QMUL-OpenLogo dataset, 
we further benchmarked a fully supervised learning setting where each logo class has real training data.
In particular, for every logo class, we made a 60\%/10\%/30\% train/val/test image split at random.
The data statistics are detailed in Table \ref{tab:fully supervise}.
With Faster-RCNN, we obtain a mAP performance of 48.3\%, 
much higher than the best corresponding open logo detection rate 13.3\% (Table \ref{tab:main}).
%


\begin{table} [h]
	\centering
	\setlength{\tabcolsep}{0.3cm}
	\caption{
		The data split statistics for the supervised learning setting on 
		QMUL-OpenLogo. 
	}
	\label{tab:fully supervise}
	\vspace{0.05cm}
	{%
		\begin{tabular}{l||c|c|c|c|c}
			\hline
			Setting & Classes & Train & Val & Test & Total   \\
			\hline\hline
			Supervised Learning
			& 352 			& 15,975 & 2,777 & 8,331 & 27,083 \\
			\hline
			
		\end{tabular}
	}
\end{table}

\section{Conclusion}

In this work, we presented a new benchmark called QMUL-OpenLogo
for enabling faithful performance test of logo detection algorithms
in more realistic and challenging deployment scenarios.
In contrast to existing closed benchmarks,
QMUL-OpenLogo considers an open-end logo detection scenario
where most classes are unsupervised --
a simulation of incrementally arriving new logo classes
without exhaustively labelled training data 
at fine-grained bounding box level during deployment.
This benchmark therefore uniquely provides a more realistic evaluation of 
algorithms for logo detection in scalable and dynamic deployment
with limited labelling budget. 
We further introduced a Context Adversarial Learning (CAL) approach
to synthetic training data generation for enabling 
the learning optimisation of state-of-the-art supervised object detection model 
even given unsupervised logo classes. 
Empirical evaluations show the performance advantages of our CAL method over the
state-of-the-art alternative detection and synthesising methods on 
the newly introduced QMUL-OpenLogo benchmark.
We also provided detailed model performance analyses on
different types of test data for giving insights on 
the specific challenges of the proposed more realistic
open logo detection.

\section*{Acknowledgement}
This work was partially supported by the China Scholarship Council, 
Vision Semantics Ltd, Royal Society Newton Advanced Fellowship Programme (NA150459),
and Innovate UK Industrial Challenge Project on Developing and Commercialising Intelligent Video Analytics Solutions for Public Safety (98111-571149).

\bibliography{record}

\begin{thebibliography}{42}
\providecommand{\natexlab}[1]{#1}
\providecommand{\url}[1]{\texttt{#1}}
\expandafter\ifx\csname urlstyle\endcsname\relax
  \providecommand{\doi}[1]{doi: #1}\else
  \providecommand{\doi}{doi: \begingroup \urlstyle{rm}\Url}\fi

\bibitem[Bianco et~al.(2017)Bianco, Buzzelli, Mazzini, and
  Schettini]{bianco2017deep}
Simone Bianco, Marco Buzzelli, Davide Mazzini, and Raimondo Schettini.
\newblock Deep learning for logo recognition.
\newblock \emph{Neurocomputing}, 245:\penalty0 23--30, 2017.

\bibitem[Boia et~al.(2014)Boia, Bandrabur, and Florea]{boia2014local}
Raluca Boia, Alessandra Bandrabur, and Catalin Florea.
\newblock Local description using multi-scale complete rank transform for
  improved logo recognition.
\newblock In \emph{IEEE International Conference on Communications}, pages
  1--4, 2014.

\bibitem[Denton et~al.(2015)Denton, Chintala, Fergus, et~al.]{denton2015deep}
Emily~L Denton, Soumith Chintala, Rob Fergus, et~al.
\newblock Deep generative image models using a￼ laplacian pyramid of
  adversarial networks.
\newblock In \emph{Advances in Neural Information Processing Systems}, pages
  1486--1494, 2015.

\bibitem[Everingham et~al.(2010)Everingham, Van~Gool, Williams, Winn, and
  Zisserman]{everingham2010pascal}
Mark Everingham, Luc Van~Gool, Christopher~KI Williams, John Winn, and Andrew
  Zisserman.
\newblock The pascal visual object classes (voc) challenge.
\newblock \emph{International journal of computer vision}, 88\penalty0
  (2):\penalty0 303--338, 2010.

\bibitem[Georgakis et~al.(2017)Georgakis, Mousavian, Berg, and
  Kosecka]{georgakis2017synthesizing}
Georgios Georgakis, Arsalan Mousavian, Alexander~C Berg, and Jana Kosecka.
\newblock Synthesizing training data for object detection in indoor scenes.
\newblock \emph{arXiv preprint arXiv:1702.07836}, 2017.

\bibitem[Girshick(2015)]{girshick2015fast}
Ross Girshick.
\newblock Fast r-cnn.
\newblock In \emph{IEEE International Conference on Computer Vision}, 2015.

\bibitem[Girshick et~al.(2014)Girshick, Donahue, Darrell, and
  Malik]{girshick2014rich}
Ross Girshick, Jeff Donahue, Trevor Darrell, and Jitendra Malik.
\newblock Rich feature hierarchies for accurate object detection and semantic
  segmentation.
\newblock In \emph{IEEE Conference on Computer Vision and Pattern Recognition},
  2014.

\bibitem[Goodfellow et~al.(2014)Goodfellow, Pouget-Abadie, Mirza, Xu,
  Warde-Farley, Ozair, Courville, and Bengio]{goodfellow2014generative}
Ian Goodfellow, Jean Pouget-Abadie, Mehdi Mirza, Bing Xu, David Warde-Farley,
  Sherjil Ozair, Aaron Courville, and Yoshua Bengio.
\newblock Generative adversarial nets.
\newblock In \emph{Advances in neural information processing systems}, pages
  2672--2680, 2014.

\bibitem[Goodfellow et~al.(2016)Goodfellow, Bengio, Courville, and
  Bengio]{goodfellow2016deep}
Ian Goodfellow, Yoshua Bengio, Aaron Courville, and Yoshua Bengio.
\newblock \emph{Deep learning}, volume~1.
\newblock MIT press Cambridge, 2016.

\bibitem[Gupta et~al.(2016)Gupta, Vedaldi, and Zisserman]{gupta2016synthetic}
Ankush Gupta, Andrea Vedaldi, and Andrew Zisserman.
\newblock Synthetic data for text localisation in natural images.
\newblock \emph{arXiv e-prints}, 2016.

\bibitem[Hoi et~al.(2015)Hoi, Wu, Liu, Wu, Wang, Xue, and Wu]{hoi2015logo}
Steven~CH Hoi, Xiongwei Wu, Hantang Liu, Yue Wu, Huiqiong Wang, Hui Xue, and
  Qiang Wu.
\newblock Logo-net: Large-scale deep logo detection and brand recognition with
  deep region-based convolutional networks.
\newblock \emph{arXiv preprint arXiv:1511.02462}, 2015.

\bibitem[Iandola et~al.(2015)Iandola, Shen, Gao, and
  Keutzer]{iandola2015deeplogo}
Forrest~N Iandola, Anting Shen, Peter Gao, and Kurt Keutzer.
\newblock Deeplogo: Hitting logo recognition with the deep neural network
  hammer.
\newblock \emph{arXiv}, 2015.

\bibitem[Isola et~al.(2017)Isola, Zhu, Zhou, and Efros]{isola2017image}
Phillip Isola, Jun-Yan Zhu, Tinghui Zhou, and Alexei~A Efros.
\newblock Image-to-image translation with conditional adversarial networks.
\newblock In \emph{IEEE Conference on Computer Vision and Pattern Recognition},
  2017.

\bibitem[Joly and Buisson(2009)]{joly2009logo}
Alexis Joly and Olivier Buisson.
\newblock Logo retrieval with a contrario visual query expansion.
\newblock In \emph{ACM International Conference on Multimedia}, pages 581--584,
  2009.

\bibitem[Kalantidis et~al.(2011)Kalantidis, Pueyo, Trevisiol, van Zwol, and
  Avrithis]{kalantidis2011scalable}
Yannis Kalantidis, Lluis~Garcia Pueyo, Michele Trevisiol, Roelof van Zwol, and
  Yannis Avrithis.
\newblock Scalable triangulation-based logo recognition.
\newblock In \emph{ACM International Conference on Multimedia Retrieval},
  page~20, 2011.

\bibitem[Kingma and Ba(2014)]{kingma2014adam}
Diederik Kingma and Jimmy Ba.
\newblock Adam: A method for stochastic optimization.
\newblock \emph{arXiv}, 2014.

\bibitem[Krizhevsky et~al.(2012)Krizhevsky, Sutskever, and
  Hinton]{krizhevsky2012imagenet}
Alex Krizhevsky, Ilya Sutskever, and Geoffrey~E Hinton.
\newblock Imagenet classification with deep convolutional neural networks.
\newblock In \emph{Advances in Neural Information Processing Systems}, pages
  1097--1105, 2012.

\bibitem[LeCun et~al.(2015)LeCun, Bengio, and Hinton]{lecun2015deep}
Yann LeCun, Yoshua Bengio, and Geoffrey Hinton.
\newblock Deep learning.
\newblock \emph{nature}, 521\penalty0 (7553):\penalty0 436, 2015.

\bibitem[Li et~al.(2014)Li, Chen, Su, Duh, Zhang, and Li]{li2014logo}
Kuo-Wei Li, Shu-Yuan Chen, Songzhi Su, Der-Jyh Duh, Hongbo Zhang, and Shaozi
  Li.
\newblock Logo detection with extendibility and discrimination.
\newblock \emph{Multimedia tools and applications}, 72\penalty0 (2):\penalty0
  1285--1310, 2014.

\bibitem[Liao et~al.(2017)Liao, Lu, Zhang, Wang, and Tang]{liao2017mutual}
Yuan Liao, Xiaoqing Lu, Chengcui Zhang, Yongtao Wang, and Zhi Tang.
\newblock Mutual enhancement for detection of multiple logos in sports videos.
\newblock In \emph{IEEE International Conference on Computer Vision}, 2017.

\bibitem[Lin et~al.(2014)Lin, Maire, Belongie, Hays, Perona, Ramanan,
  Doll{\'a}r, and Zitnick]{lin2014microsoft}
Tsung-Yi Lin, Michael Maire, Serge Belongie, James Hays, Pietro Perona, Deva
  Ramanan, Piotr Doll{\'a}r, and C~Lawrence Zitnick.
\newblock Microsoft coco: Common objects in context.
\newblock In \emph{European Conference on Computer Vision}. 2014.

\bibitem[Liu et~al.(2016)Liu, Anguelov, Erhan, Szegedy, and Reed]{liu2015ssd}
Wei Liu, Dragomir Anguelov, Dumitru Erhan, Christian Szegedy, and Scott Reed.
\newblock Ssd: Single shot multibox detector.
\newblock In \emph{European Conference on Computer Vision}, 2016.

\bibitem[Mathieu et~al.(2015)Mathieu, Couprie, and LeCun]{mathieu2015deep}
Michael Mathieu, Camille Couprie, and Yann LeCun.
\newblock Deep multi-scale video prediction beyond mean square error.
\newblock \emph{arXiv preprint arXiv:1511.05440}, 2015.

\bibitem[Pan et~al.(2013)Pan, Yan, Xu, Sun, Shao, and Wu]{pan2013vehicle}
Chun Pan, Zhiguo Yan, Xiaoming Xu, Mingxia Sun, Jie Shao, and Di~Wu.
\newblock Vehicle logo recognition based on deep learning architecture in video
  surveillance for intelligent traffic system.
\newblock In \emph{IET International Conference on Smart and Sustainable City},
  pages 123--126, 2013.

\bibitem[Pathak et~al.(2016)Pathak, Krahenbuhl, Donahue, Darrell, and
  Efros]{pathak2016context}
Deepak Pathak, Philipp Krahenbuhl, Jeff Donahue, Trevor Darrell, and Alexei~A
  Efros.
\newblock Context encoders: Feature learning by inpainting.
\newblock In \emph{IEEE Conference on Computer Vision and Pattern Recognition},
  pages 2536--2544, 2016.

\bibitem[Peng et~al.(2015)Peng, Sun, Ali, and Saenko]{peng2015learning}
Xingchao Peng, Baochen Sun, Karim Ali, and Kate Saenko.
\newblock Learning deep object detectors from 3d models.
\newblock In \emph{IEEE International Conference on Computer Vision}, pages
  1278--1286, 2015.

\bibitem[Radford et~al.(2015)Radford, Metz, and
  Chintala]{radford2015unsupervised}
Alec Radford, Luke Metz, and Soumith Chintala.
\newblock Unsupervised representation learning with deep convolutional
  generative adversarial networks.
\newblock \emph{arXiv preprint arXiv:1511.06434}, 2015.

\bibitem[Redmon and Farhadi(2017)]{redmon2017yolo9000}
Joseph Redmon and Ali Farhadi.
\newblock Yolo9000: better, faster, stronger.
\newblock In \emph{IEEE Conference on Computer Vision and Pattern Recognition},
  2017.

\bibitem[Redmon et~al.(2016)Redmon, Divvala, Girshick, and
  Farhadi]{redmon2015you}
Joseph Redmon, Santosh Divvala, Ross Girshick, and Ali Farhadi.
\newblock You only look once: Unified, real-time object detection.
\newblock In \emph{IEEE Conference on Computer Vision and Pattern Recognition},
  2016.

\bibitem[Ren et~al.(2015)Ren, He, Girshick, and Sun]{ren2015faster}
Shaoqing Ren, Kaiming He, Ross Girshick, and Jian Sun.
\newblock Faster r-cnn: Towards real-time object detection with region proposal
  networks.
\newblock In \emph{Advances in Neural Information Processing Systems}, pages
  91--99, 2015.

\bibitem[Revaud et~al.(2012)Revaud, Douze, and Schmid]{revaud2012correlation}
Jerome Revaud, Matthijs Douze, and Cordelia Schmid.
\newblock Correlation-based burstiness for logo retrieval.
\newblock In \emph{ACM International Conference on Multimedia}, pages 965--968,
  2012.

\bibitem[Romberg and Lienhart(2013)]{romberg2013bundle}
Stefan Romberg and Rainer Lienhart.
\newblock Bundle min-hashing for logo recognition.
\newblock In \emph{Proceedings of the 3rd ACM conference on International
  conference on multimedia retrieval}, pages 113--120. ACM, 2013.

\bibitem[Romberg et~al.(2011)Romberg, Pueyo, Lienhart, and
  Van~Zwol]{romberg2011scalable}
Stefan Romberg, Lluis~Garcia Pueyo, Rainer Lienhart, and Roelof Van~Zwol.
\newblock Scalable logo recognition in real-world images.
\newblock In \emph{Proceedings of the 1st ACM International Conference on
  Multimedia Retrieval}, page~25. ACM, 2011.

\bibitem[Russakovsky et~al.(2015)Russakovsky, Deng, Su, Krause, Satheesh, Ma,
  Huang, Karpathy, Khosla, Bernstein, et~al.]{russakovsky2015imagenet}
Olga Russakovsky, Jia Deng, Hao Su, Jonathan Krause, Sanjeev Satheesh, Sean Ma,
  Zhiheng Huang, Andrej Karpathy, Aditya Khosla, Michael Bernstein, et~al.
\newblock Imagenet large scale visual recognition challenge.
\newblock \emph{International Journal of Computer Vision}, 115\penalty0
  (3):\penalty0 211--252, 2015.

\bibitem[Salimans et~al.(2016)Salimans, Goodfellow, Zaremba, Cheung, Radford,
  and Chen]{salimans2016improved}
Tim Salimans, Ian Goodfellow, Wojciech Zaremba, Vicki Cheung, Alec Radford, and
  Xi~Chen.
\newblock Improved techniques for training gans.
\newblock In \emph{Advances in Neural Information Processing Systems}, pages
  2234--2242, 2016.

\bibitem[Su et~al.(2017{\natexlab{a}})Su, Gong, and Zhu]{su2017weblogo}
Hang Su, Shaogang Gong, and Xiatian Zhu.
\newblock Weblogo-2m: Scalable logo detection by deep learning from the web.
\newblock In \emph{Workshop of the IEEE International Conference on Computer
  Vision}, 2017{\natexlab{a}}.

\bibitem[Su et~al.(2017{\natexlab{b}})Su, Zhu, and Gong]{su2016deep}
Hang Su, Xiatian Zhu, and Shaogang Gong.
\newblock Deep learning logo detection with data expansion by synthesising
  context.
\newblock \emph{IEEE Winter Conference on Applications of Computer Vision},
  2017{\natexlab{b}}.

\bibitem[Su et~al.(2018)Su, Gong, and Zhu]{su2018scalable}
Hang Su, Shaogang Gong, and Xiatian Zhu.
\newblock Scalable deep learning logo detection.
\newblock \emph{arXiv preprint arXiv:1803.11417}, 2018.

\bibitem[Su et~al.(2012)Su, Deng, and Fei-Fei]{su2012crowdsourcing}
Hao Su, Jia Deng, and Li~Fei-Fei.
\newblock Crowdsourcing annotations for visual object detection.
\newblock In \emph{Workshops at the Twenty-Sixth AAAI Conference on Artificial
  Intelligence}, volume~1, 2012.

\bibitem[Su et~al.(2015)Su, Qi, Li, and Guibas]{su2015render}
Hao Su, Charles~R Qi, Yangyan Li, and Leonidas~J Guibas.
\newblock Render for cnn: Viewpoint estimation in images using cnns trained
  with rendered 3d model views.
\newblock In \emph{IEEE International Conference on Computer Vision}, 2015.

\bibitem[T{\"u}zk{\"o} et~al.(2017)T{\"u}zk{\"o}, Herrmann, Manger, and
  Beyerer]{tuzko2017open}
Andras T{\"u}zk{\"o}, Christian Herrmann, Daniel Manger, and J{\"u}rgen
  Beyerer.
\newblock Open set logo detection and retrieval.
\newblock \emph{arXiv preprint arXiv:1710.10891}, 2017.

\bibitem[Wang and Gupta(2016)]{wang2016generative}
Xiaolong Wang and Abhinav Gupta.
\newblock Generative image modeling using style and structure adversarial
  networks.
\newblock In \emph{European Conference on Computer Vision}, pages 318--335,
  2016.

\end{thebibliography}
\end{document}